# Metric Classification Network in Actual Face Recognition Scene

Jian Li[1,2], Yan Wang[1,2], Xiubao Zhang[2], Weihong Deng[1], Haifeng Shen[2]

[1]Beijing University of Posts and Telecommunications, Beijing 100876, China
[2]AI Labs, Didi Chuxing, Beijing 100193, China
lijian931212@126.com, wang_y@bupt.edu.cn,
zhangxiubao@didiglobal.com, whdeng@bupt.edu.cn, shenhaifeng@didiglobal.com

**Abstract**

In order to make facial features more discriminative, some new models have recently been proposed. However, almost all of these models use the traditional face verification method, where the cosine operation is performed using the features of the bottleneck layer output. However, each of these models needs to change a threshold each time it is operated on a different test set. This is very inappropriate for application in real-world scenarios. In this paper, we train a validation classifier to normalize the decision threshold, which means that the result can be obtained directly without replacing the threshold. We refer to our model as validation classifier, which achieves best result on the structure consisting of one convolution layer and six fully connected layers. To test our approach, we conduct extensive experiments on Labeled Face in the Wild (LFW) and Youtube Faces (YTF), and the relative error reduction is 25.37% and 26.60% than traditional method respectively. These experiments confirm the effectiveness of validation classifier on face recognition task.

## Introduction

Face recognition, as one of the most common computer vision tasks, usually includes two subtasks: face verification and face identification. Face verification compares a pair of faces to decide whether they derive from the same identity, while face identification classifies a face to a specific identity. Over the past few years, owing to the more and more excellent network architectures (Krizhevsky et al. 2012; Simonyan et al. 2014; Szegedy et al. 2015; He et al. 2015) and more and more masterly learning tricks (Sun et al. 2014; Schroff et al. 2015; Wen et al. 2016), convolutional neural networks (CNNs) have significantly improved the performance of face recognition (Schroff et al. 2015; Wen et al. 2016; Liu et al. 2017; Wang et al. 2018; Deng et al. 2018; Wang et al. 2018). Although excellent performance of face verification on some public and authoritative datasets such as Labeled Faces in the Wild (LFW), there are still a lot of difficulties in face recognition that need to be overcome, especially when applying face recognition technology in practice.

Intuitively, to get better performance, the learned features ought to be more discriminative. That is to say, the intra-class distance needs to be compact and simultaneously the inter-class separability needs to be maximized. Recent works mostly follow the principle and focus on creating novel loss functions such as SphereFace (Liu et al. 2017). Through bringing in margin constraint in training, the classification network becomes more stringent and the learned features are more discriminative. However, SphereFace, like most networks, uses traditional face verification method to judge whether a pair of images come from the same identity.

The face verification method in the traditional process is generally: input two face images to be verified into the face recognition model, obtain two feature vectors, calculate the cosine of the angle between them, and compare with the threshold to judge whether they are from the same identity. The accuracy of the test set is measured by a 10-fold cross validation procedure and can be summarized as: the feature mean is calculated by the other nine folds, and the optimal threshold is the threshold that maximizes the accuracy of the other nine folds. Then, the feature vector obtained by the tenth fold subtracts the feature mean, and then the cosine is calculated and compared with the optimal threshold. However, in practical applications, the face verification system will be applied to multiple test sets. The optimal threshold for each test set is different and varies greatly, which means that the threshold will be changed whenever a different test set is tested. In this paper, we train the validation classifier to normalize the decision threshold, that is, directly obtain the result without replacing the optimal threshold. The details are described in Section 3.

Our contributions can be summarized as follows: we train a neural network, called verification classifier, to replace the traditional cosine similarity metric method to address the problem of changing the optimal threshold for each test set tested. We jointly train the face recognition model and the verification classifier in an end-to-end manner. We introduce a focusing factor into the loss function that can make the training to focus on the hard sample pairs online, which can ensure increasing hard sample pairs as the network trains. Trained on CASIA-WebFace dataset, the verification classifier achieves better results on several datasets such as Labeled Face in the Wild (LFW) and Youtube Faces (YTF).

## Related work

Recently, many approaches have been proposed to improve the face verification performance, and these methods can be roughly divided into two categories. One type Treats face recognition as a classification task, and every identity is taken as a class (Liu et al. 2017; Wang et al. 2017; Liu et al. 2016), which improve face verification performance by optimizing the intermediate bottleneck layer. The other type employs metric learning as the measure of optimizing, such as contrastive loss (Sun et al. 2014; Chopra et al. 2005) and triplet loss (Schroff et al. 2015), which learn a mapping from face images to a compact Euclidean space where distances directly correspond to a measure of face similarity. They use a deep convolutional network trained to directly optimize the embedding itself rather than an intermediate bottleneck layer as in the former deep learning approaches. However, whether it is the first method or the second method, they need to perform the cosine metrics on the extracted features and compare it with the selected thresholds. Therefore, the above problems will still occur in practical applications. Sun et al. 2014 and Han et al. 2015 adopt the method of training a neural network that computes a similarity between the extracted features for face verification. However, both the architectures of the two neural networks are wide and shallow, which have an average performance. Since the research on metric learning modeling the similarity between features using metric neural network has not been deepened, we have carried out new research on this method. The overall process is: input the concatenation of a pair of features to the network and output two values in [0; 1] from the metric neural network, these are non-negative, sum up to one, and can be interpreted as the network's estimate of the two features match and do not match, respectively.

In comparison, our proposed feature extraction method uses a deep convolutional network with multiple convolutional and spatial pooling layers plus an optional bottle neck layer to obtain feature vectors, followed by a similarity measure also based on neural network. We use a neural network to learn the pairwise similarity, which has the potential to embrace more complex similarity functions beyond distance metrics such as Euclidean. Our method is complementary to existing face verification similarity measure. Different from these metric learning methods, our proposed method adopts convolutional neural network and learns a nonlinear transformation to project face pairs into one feature space in a deep architecture. We also achieve the very competitive performance on the face verification in the wild problem with several existing publicly available datasets.

In this paper, we propose a novel model named verification classifier. There are two main contributions. The first contribution is that it is an effective alternative to Euclidean distance and cosine similarity in metric learning problem. The second contribution is that the verification classifier can improve the generalization ability of an existing metric significantly in most cases. Our method is different from all the above methods in terms of distance measures. All of the above methods use 10-fold cross validation procedure or simple neural network whilst our method uses complex neural network which leads to a simple and effective metric learning method. The rest of this paper is structured as follows. Section 3 present how verification classifier can be applied to face verification. Experimental results are presented in section 4. Finally, conclusion is given in section 5.

## Verification classifier

In this section, we will first describe how we extract features. Then introduce how the network structure we use is chosen. Finally, we describe how we construct training sets and perform hard sample mining.

### Feature optimization

As we said earlier, many methods have been proposed to improve facial verification performance, and these methods can be broadly classified into two categories. Since the verification classifier we are going to learn in this paper uses the neural network for metric learning, we use the first method to extract features by optimizing the intermediate bottleneck layer to improve the performance. We use the method of Sphereface to optimize features.

We start with the softmax loss and then introduce A-Softmax loss. The softmax loss can be written as

$$L = \frac{1}{N}\sum_i -log(\frac{e^{f_{y_i}}}{\sum_j e^{f_j}}) \qquad (3)$$

Taking the case of binary-class example, the posterior probabilities of softmax can be written:

$$p_1 = \frac{exp(W_1^T x + b_1)}{exp(W_1^T x + b_1) + exp(W_2^T x + b_2)} \qquad (4)$$

$$p_2 = \frac{exp(W_2^T x + b_2)}{exp(W_1^T x + b_1) + exp(W_2^T x + b_2)} \qquad (5)$$

where $x$ represents the feature. $W_i$ and $b_i$ represent the weights and biases of the last fully connected layer. If $p_1 > p_2$, the estimated tag will be classified to category 1. If $p_1 < p_2$, the estimated tag will be classified to category 2. By comparing $p_1$ and $p_2$, it is clear that $W_1^T x + b_1$ and $W_2^T x + b_2$ determine the classification result. After rewriting $W_i^T x + b_i$ to $\|W_i^T\| \|x\| cos(\theta_i) + b_i$ where $\theta_i$ is the angle between $W_i$ and $x$, normalizing the weights and making the biases be zero, the posterior probabilities become $p_1 = \|x\| cos(\theta_1)$ and $p_2 = \|x\| cos(\theta_2)$. Since $p_1$ and $p_2$ share the same $x$, the consequence depends only on the values of $\theta_1$ and $\theta_2$. Assuming that the feature $x$ from class 1 is given, it is known that $cos(\theta_1) > cos(\theta_2)$ is required to classify $x$ correctly. Then change the requirement to $cos(m\theta_1) > cos(\theta_2)$, where $m$ is a coefficient in order to more accurately classify $x$. It basically makes the decision more stringent. The above analysis built on a binary-class case can be generalized to a multi-class case trivially. So, the equation (3) can be formulated as:

$$L_A = \frac{1}{N}\sum_i -log\left(\frac{exp(\|x_i\|\varphi(\theta_{y_i,i}))}{exp(\|x_i\|\varphi(\theta_{y_i,i}))+\sum_{j\neq y_i} exp(\|x_i\|cos(\theta_{j,i}))}\right) \quad (6)$$

where we define: $\varphi(\theta_{y_i,i}) = (-1)^k cos(m\theta_{y_i,i}) - 2k$, $\theta_{y_i,i} \in [\frac{k\pi}{m}, \frac{(k+1)\pi}{m}]$ and $k \in [0, m-1]$. $m \geq 1$ is an integer that controls the size of angular margin.

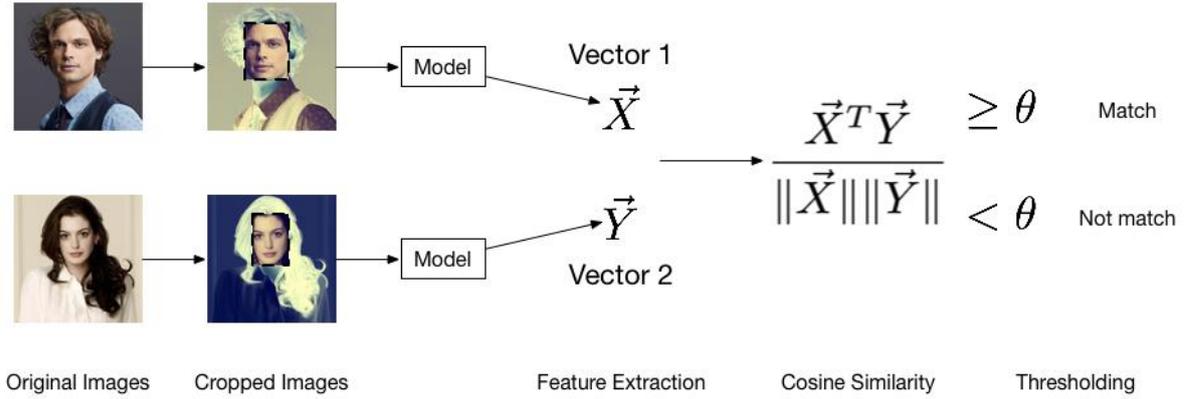

Figure 1

**Selection of Network Structure**

The overview of traditional method (Nguyen et al. 2010) is presented in Figure 1. First, two original images are cropped to smaller sizes. Next some feature extraction method is used to form feature vectors from the cropped images. Cosine similarity between the vectors is the similarity score between two faces. Finally, the score is compared to the threshold to determine if the two faces are the same. The optimal threshold θ is estimated from the training set. Specifically, θ is set so that False Acceptance Rate equals to False Rejection Rate. As mentioned above, this method is not suitable for practical application. Therefore, we introduce neural networks to calculate similarities between features. In contrast, here we aim for a similarity function that can account for a broader set of appearance changes and can be used in a much wider and more challenging set of applications.

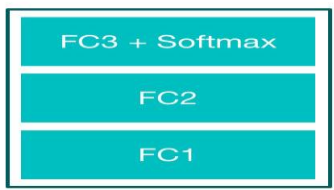

Figure 2

Han et al. 2015 is a measure of similarity using a neural network, but its network structure has only three layers, as shown in Figure 2. Intuitively, such a shallow network structure does not achieve the best accuracy. So on this basis, we use more layers as our network structure, we will explore the impact of layer number on verifying the accuracy of the classifier in the fourth section.

In addition to the network structure mentioned above, we also reference two network structures for comparing the similarity functions between image blocks (Zagoruyko et al. 2015), to which the input is considered to be a pair of image patches. There are two main types of network structures for comparing the similarity functions between image blocks: Siamese networks and 2-channel networks.

Siamese: as shown in Figure 3, this type of network resembles the idea of having a descriptor. There are two branches in the network that share exactly the same architecture and the same set of weights. Each branch takes as input one of the two images and then applies a series of convolutional, ReLU and max-pooling layers. Branch outputs are concatenated and given to a top network that consists of linear fully connected and ReLU layers. Zagoruyko et al. 2015 uses a top network consisting of 2 linear fully connected layers (each with 512 hidden units) that are separated by a ReLU activation layer. Branches of the siamese network can be viewed as descriptor computation modules and the top network can be viewed as a similarity function. For the task of matching two sets of images at test time, descriptors can first be computed independently using the branches and then matched with the top network (or even with a distance function like L2).

2-channel: as shown in Figure 4, unlike the previous models, here there is no direct notion of descriptor in the architecture. We simply consider the two images of an input pair as a 2-channel image, which is directly fed to the first convolutional layer of the network. In this case, the bottom part of the network consists of a series of convolutional, ReLU and max-pooling layers. The output of this part is then given as input to a top module that consists simply of a fully connected linear decision layer with one output. This network provides greater flexibility compared to the above models as it starts by processing the two patches jointly. Furthermore, it is fast to train.

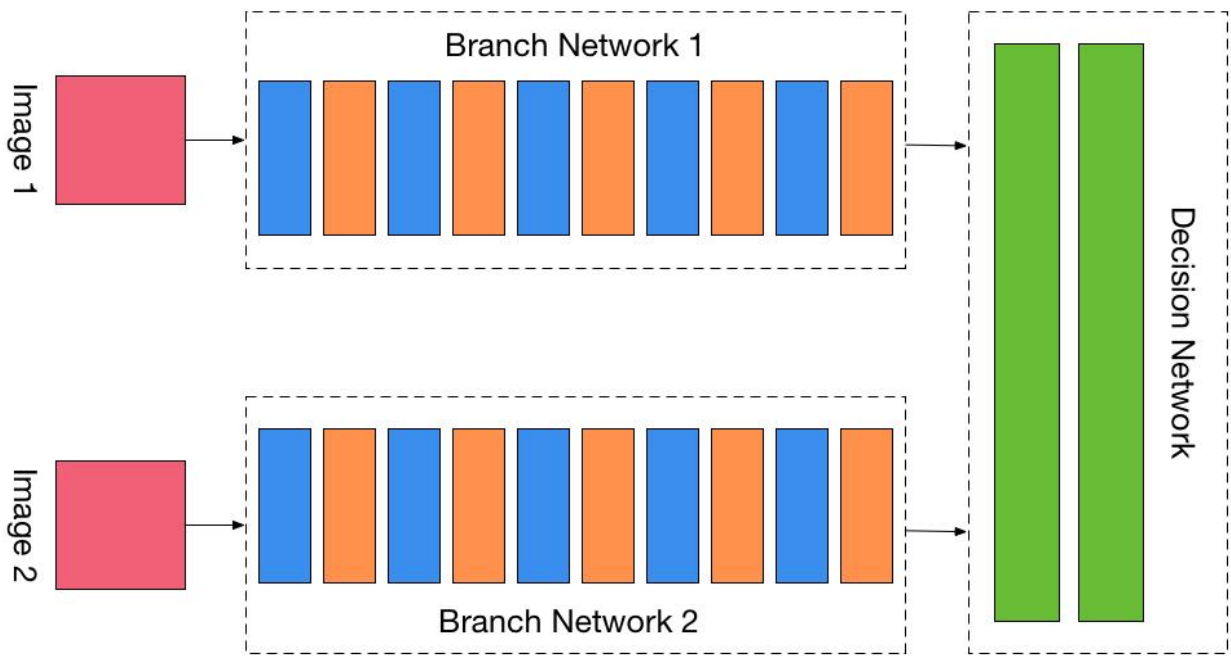

Figure 3

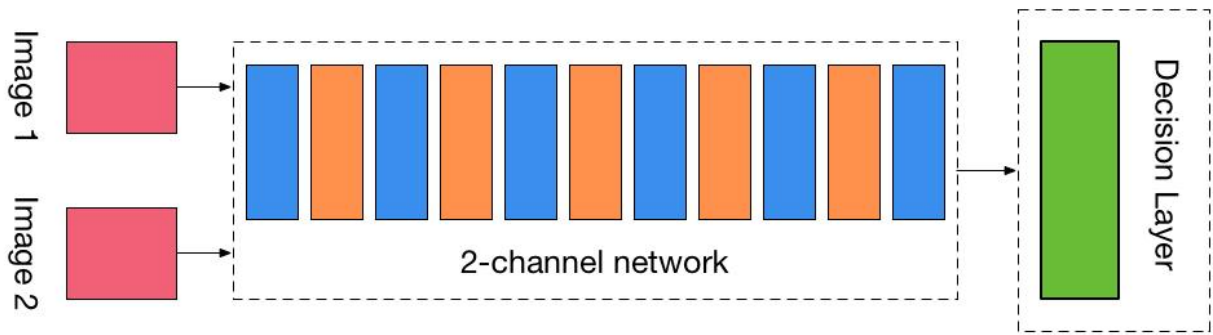

Figure 4

Essentially these architectures stem from the different way that each of them attempts to address the following question: when composing a similarity function for comparing image patches, do we first choose to compute a descriptor for each patch and then create a similarity on top of these descriptors or do we perhaps choose to skip the part related to the descriptor computation and directly proceed with the similarity estimation? Since we measure the similarity of feature vector pairs rather than pairs of pictures, we use the latter to directly estimate the similarity. So, we explored and studied the 2-channel neural network architecture, which is specifically suitable for this task. We improve Figure 4 to get the network structure used in our experiments, as shown in Figure 5. We show that this approach can be significantly better than the latest technology on several issues and benchmark data sets. In the 2-channel network, two features are stacked (i.e. each feature serves as a channel for the composite, paired feature) as the input to a neural network. The neural network then leads to a full connected linear decision layer with 2 outputs that indicate the similarity of the two features. Further details about our adaptation are in the results section.

Since we are comparing feature pairs rather than image pairs, we can't design the network structure as a convolutional neural network like Figure 3, but we can't use only the full connection layer as in Figure 5. Because if only the fully connected layer is used as shown in Figure 5, the 2-channel network structure is no different from the single channel network structure. The only difference is that the feature that is fed into the 2-channel network structure is that the two feature vectors are connected in parallel, and the feature that is fed into the single-channel network structure is that the two feature vectors are connected in series. If so, the 2-channel network will lose its effect, that is, not combining the two features. Therefore, we design the first layer of the network as a convolutional layer to fuse the two features, and the remaining layers use fully connected layers.

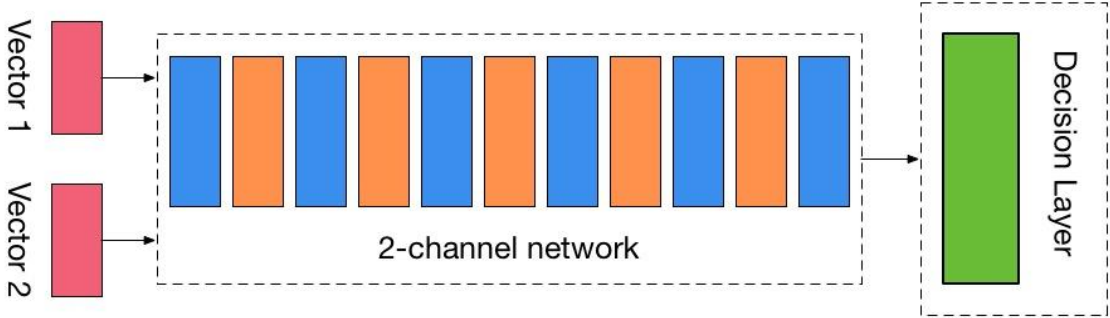

Figure 5

**Sample Pair Selection Strategy**

Sampling is important in training because positive and negative sample pairs are very unbalanced. We use the sampler to generate the same number of positive and negative pairs in each small batch so that the network does not overly bias towards negative decisions. The sampler also enforces various operations to prevent overfitting to a finite negative set. If the batch size is 32, we provide 16 positive sample pairs and 16 negative sample pairs for SGD in each training iteration. A positive sample pair is obtained by reading the next random 16 identities from the database and randomly picking two images in each identity as a pair. Since we traverse the entire data set multiple times, even if we only select one pair from each group in each pass, the network still has good positive coverage, while 16 negatives were obtained by sampling two images from different identities from the database. However, after controlling the balance of the positive and negative samples, hard sample mining is also required. Because we have about 500,000 images, after pairing, we have a total of 250 billion feature pairs combined, and it is unrealistic to train with all sample pairs. Therefore, we use random matching in training and set the sample pair threshold. For negative sample pairs, when the distance between two sample features is less than the threshold, we determine that it is a negative hard sample. For the same reason, the same is true for the selection of hard samples.

In fact, a better way to solve the positive and negative sample balance and hard sample mining is to introduce the idea of focus loss, introduce balance factor and focus factor in the loss layer, and solve the above two problems. In binary-class case, the focal loss can be formulated as:

$$L_{FL} = -\alpha_t(1-p_t)^\gamma log(p_t) \quad (8)$$

in which we define:

$$p_t = \begin{cases} p & if \ y = 1 \\ 1-p & else \end{cases} \quad (9)$$

$\alpha_t$ is defined analogously to how we defined $p_t$. In the equation (9), $y \in \{\pm 1\}$ assigns the ground truth and $p \in [0,1]$ is the predicted probability of the model for the class with label $y = 1$. Extending the focal loss (Lin et al. 2017) to many types of cases is very simple and works well.

The most frequently used method for solving class imbalance is to introduce a weighting factor $\alpha \in [0,1]$ for one class and $1-\alpha$ for the other class. But how does the focusing factor $(1-p)^\gamma$ lessen the losses of simple samples and make training focus on difficult samples? When a sample is misclassified, $1-p$ will be large, that is, the focusing factor will be close to 1, which means that the loss is hardly affected. When the focusing factor is close to 0, the loss of the well-classified example will be greatly reduced. Therefore, the adjustment factor lessens the loss contribution caused by the simple examples and makes the training pay more attention to the hard samples. As can be seen from Table V and Table VI, the focal loss can improve the final accuracy to some extent.

## Experiment

In this section, we will first describe the experimental setup. Then we will compare the differences between different network structures. Finally, we will demonstrate the superiority of our proposed method over traditional methods on the LFW (Hu et al. 2014) and YTF (Wolf et al. 2011) data sets. The LFW dataset contains 13,233 facial images belonging to 5,749 different individuals. The dataset contains a large amount of face images. The performance of validation classifier is evaluated on the LFW 6000 pairs of faces. The YTF dataset contains 3,424 videos belonging to 1,595 different identities, which contains a large amount of face images.

### Experiment Settings

We train our CNN model using the publicly available network-collected dataset CASIA (Yi et al. 2014), which contains 490K face images from 10,575 individuals. All face key points in the images are detected by the MTCNN

(Zhang et al. 2016). We use five facial landmarks for similarity transformation to normalize the face images. Each pixel in the image is normalized. To be fair, all methods use the same CNN structure. We evaluate our approach on the 64-layer CNN architecture given in Table I, which shows the specific settings for the different CNNs we use. The CNN architecture contains convolution units, residual units and fully connected units. We set the batch size as 256 and distribute the mini-batch on 2 GPUs.

Table I  CNN architectures

| Layer | 64-layer CNN |
|---|---|
| Conv1.x | $[3 \times 3, 64] \times 1, S2$ <br> $\begin{bmatrix} 3 \times 3, 64 \\ 3 \times 3, 64 \end{bmatrix} \times 3$ |
| Conv2.x | $[3 \times 3, 128] \times 1, S2$ <br> $\begin{bmatrix} 3 \times 3, 128 \\ 3 \times 3, 128 \end{bmatrix} \times 8$ |
| Conv3.x | $[3 \times 3, 256] \times 1, S2$ <br> $\begin{bmatrix} 3 \times 3, 256 \\ 3 \times 3, 256 \end{bmatrix} \times 16$ |
| Conv4.x | $[3 \times 3, 512] \times 1, S2$ <br> $\begin{bmatrix} 3 \times 3, 512 \\ 3 \times 3, 512 \end{bmatrix} \times 3$ |
| B | 512 |
| F | 1024 |
| VC | $2 \times 1024$ |
| VC_CONV | $[1 \times 1, 256] \times 1$ |
| VC_FC.x | $[FC] \times 5$ |
| FC | 2 |

In the Table I, B represents the bottleneck layer dimension, F represents the combination of the bottleneck feature of the picture and the flipped picture, VC represents the dimension of the F features of the two pictures to be verified in parallel, and VC_CONV represents the first convolutional layer of the verification classification. VC_FC.x represents the next few fully connected layers. FC stands for the classification layer of the verification classifier.

**Exploratory Experiments**

In order to study the impact of network structure on accuracy, we conduct comparative experiments on different structures based on Figure 5. We have mainly studied the number of network layers and the types of layers. All experiments are based on a network structure with only fully connected layers. Each fully connected layer is followed by a dropout layer. It can be seen from the experimental results in Table II, that the network structure of the 7-layer fully connected layer has the best effect.

Table II Error rate (%) and Relative error reduction (%) on LFW

| Layers | Error rate | Relative error reduction |
|---|---|---|
| 3 | 1.03 | -- |
| 5 | 0.83 | 22.42 |
| **7** | **0.62** | **39.81** |
| 9 | 1.27 | -20.30 |

After that we all experimented on the 7-layer structure. We changed the first fully connected layer to the convolution layer and experimented again, and added the BN layer for comparison experiments. The results are shown in Table III.

Table III Error rate (%) and Relative error reduction (%) on LFW

| 7-Layers | Error rate | Relative error reduction |
|---|---|---|
| 7-FC | 0.62 | -- |
| 7-FC+BN | 0.68 | -9.68 |
| **1-CONV+6-FC** | **0.55** | **11.29** |

From the experimental results, it can be seen that after using the BN layer instead of the dropout layer after the fully connected layer, the accuracy of the result is reduced. But after replacing the first fully connected layer with a convolutional layer, the experimental result improves a lot.

But we don't know how to set the number of kernels of the convolution layer to maximize the experimental result after replacing the first fully connected layer with the convolutional layer. Therefore, we use a different number of convolution kernels to do a series of comparative experiments. The experimental results are shown in Table IV.

Table IV Error rate (%) and Relative error reduction (%) on LFW

| Kernels | Error rate | Relative error reduction |
|---|---|---|
| 64 | 0.55 | -- |
| **128** | **0.52** | **5.45** |
| 256 | 0.55 | 0.00 |

From the experimental results, we can achieve better performance when we choose k=128.

**Experiments on LFW and YTF**

We evaluate our method on Labeled Face in the Wild (LFW) firstly. To be fair, all methods use the same 64-layer CNN architecture, shown in Table I. The final results are given in Table V, from which we can see that compared with other models, the verification classifier is still very competitive. On LFW dataset, the error rate has reduced by 22.39% relatively. When the focal loss was introduced during model training, we found that the error rate dropped further to 25.37% relatively.

Table V Error rate (%) and Relative error reduction (%) on LFW

| Method | Error rate | Relative error reduction |
|---|---|---|
| COS | 0.67 | -- |
| Ours(no focal) | 0.52 | 22.39 |
| **Ours(focal)** | **0.50** | **25.37** |

We also evaluate our method on Youtube Faces (YTF). The experiments use the same 64-layer CNN architecture mentioned above. The final results are given in Table VI, from which we can see that the error rate of YTF data set has reduced by 25.60% relatively. So, we can see that the verification classifier has strong generalization ability on other datasets. When the focal loss was introduced during model training, we found that the error rate reduced further to 26.60% relatively.

**Table VI** Error rate (%) and Relative error reduction (%) on YTF

| Method | Error rate | Relative error reduction |
|---|---|---|
| COS | 5.00 | -- |
| Ours(no focal) | 3.72 | 25.60 |
| **Ours(focal)** | **3.67** | **26.60** |

## Conclusions

Conventional face verification methods use the features of the bottleneck layer output to calculate cosine similarity. However, each model needs to change the threshold each time it is operated on the test set. This is very inappropriate for applications in real-world scenarios. In this paper, we train the validation classifier to normalize the decision threshold, which means that the results can be obtained directly without having to replace the threshold. We call this model a verification classifier. We conducted a series of exploratory experiments on the different structures of the verification classifier. We mainly studied the influence of the number of network layers and the type of layers on accuracy. We find that when the network structure is seven layers and the convolutional layer is added to the network structure, the error rate has reduced by 11.29% relatively on LFW dataset. In addition, we find the relative error reduction is 25.37% and 26.60% than traditional method respectively on Labeled Face in the Wild (LFW) and Youtube Faces (YTF) when a focusing factor is introduced into the loss function.

## Acknowledgements

This work is sponsored by Didi-Chuxing research collaboration plan.